\documentclass{midl} % Include author names
\usepackage{mwe} % to get dummy images
\usepackage[utf8]{inputenc} % allow utf-8 input
\usepackage[T1]{fontenc}    % use 8-bit T1 fonts
\usepackage{url}            % simple URL typesetting
\usepackage{booktabs}       % professional-quality tables
\usepackage{amsfonts}       % blackboard math symbols
\usepackage{nicefrac}       % compact symbols for 1/2, etc.
\usepackage{microtype}      % microtypography
\usepackage{graphicx}
\usepackage{float}
\usepackage{amsmath}

\usepackage{subcaption}
\usepackage{array}
\usepackage{multirow}

\jmlrvolume{-- XXXX}
\jmlryear{2019}
\jmlrworkshop{Full Paper -- MIDL 2019}

\title[Segmenting Potentially Cancerous Areas in Prostate Biopsies]{Segmenting Potentially Cancerous Areas in Prostate Biopsies using Semi-Automatically Annotated Data}
\midlauthor{\Name{Nikolay Burlutskiy\nametag{$^{1}$}\midljointauthortext{Corresponding authors.}}               \Email{nikolay.burlutsky@contextvision.se}\\
            \Name{Nicolas Pinchaud\nametag{$^{1}$}\midlotherjointauthor}
            \Email{nicolas.pinchaud@contextvision.se}\\
            \Name{Feng Gu\nametag{$^{1}$}} 
            \Email{feng.gu@contextvision.se}\\
            \Name{Daniel Hägg\nametag{$^{1}$}} 
            \Email{daniel.hagg@contextvision.se}\\
            \Name{Mats Andersson\nametag{$^{1}$}} \Email{mats.andersson@contextvision.se}\\
            \Name{Lars Björk\nametag{$^{1,2}$}} 
            \Email{lars.bjork@contextvision.se}\\
            \Name{Kristian Eurén\nametag{$^{1}$}} \Email{kristian.euren@contextvision.se}\\
            \Name{Cristina Svensson\nametag{$^{1}$}} \Email{cristina.svensson@contextvision.se}\\
            \Name{Lena Kajland Wilén\nametag{$^{1}$}} \Email{lena.kw@contextvision.se}\\
            \Name{Martin Hedlund\nametag{$^{1}$}} \Email{martin.hedlund@contextvision.se}\\
            \addr $^{1}$ ContextVision AB, Stockholm, Sweden \\
            \addr $^{2}$ NORDFERTIL Research Lab Stockholm, Department of Women’s and Children’s Health, Karolinska Institutet and University Hospital \\
}
\begin{document}

\maketitle

\begin{abstract}
Gleason grading specified in ISUP 2014 is the clinical standard in staging prostate cancer and the most important part of the treatment decision. However, the grading is subjective and suffers from high intra and inter-user variability. To improve the consistency and objectivity in the grading, we introduced glandular tissue WithOut Basal cells (WOB) as the ground truth. The presence of basal cells is the most accepted biomarker for benign glandular tissue and the absence of basal cells is a strong indicator of acinar prostatic adenocarcinoma, the most common form of prostate cancer. Glandular tissue can objectively be assessed as WOB or not WOB by using specific immunostaining for glandular tissue (Cytokeratin 8/18) and for basal cells (Cytokeratin 5/6 + p63). Even more, WOB allowed us to develop a semi-automated data generation pipeline to speed up the tremendously time consuming and expensive process of annotating whole slide images by pathologists. We generated 295 prostatectomy images exhaustively annotated with WOB. Then we used our Deep Learning Framework, which achieved the $2^{nd}$ best reported score in Camelyon17 Challenge, to train networks for segmenting WOB in needle biopsies. Evaluation of the model on 63 needle biopsies showed promising results which were improved further by finetuning the model on 118 biopsies annotated with WOB, achieving F1-score of 0.80 and Precision-Recall AUC of 0.89 at the pixel-level. Then we compared the performance of the model against 17 biopsies annotated independently by 3 pathologists using only H\&E staining. The comparison demonstrated that the model performed on a par with the pathologists. Finally, the model detected and accurately outlined existing WOB areas in two biopsies incorrectly annotated as totally WOB-free biopsies by three pathologists and in one biopsy by two pathologists.
\end{abstract}

\section{Introduction}

The Gleason system was developed in 1966 where architectural patterns are taken as a basis for determining the score \cite{Gleason77}. The system has been revised several times to reflect tumor biology better \cite{Chen2016}. Currently, ISUP 2014 is the clinical standard in staging prostate cancer and the most important part of the treatment decision. However, the grading suffers from high intra- and inter-user variability, with reported Gleason score discordance ranging from 30-53\% \cite{google_paper18}.

The goal of this research is to develop a decision support tool to aid pathologists in their work. We trained Deep Learning (DL) models to segment potentially cancerous areas in prostate biopsies. The models will provide relevant regions of slides for pathologists to focus on, helping them to work more efficiently and save time. Unfortunately, it is difficult to train a DL network on Gleason annotations since it is almost impossible to acquire a consistent Ground Truth for comparison. As a first step in prostate cancer grading, we introduced a class that we objectively have more control of, Glandular tissue WithOut Basal cells (WOB) class. The presence of basal cells is the most accepted biomarker for benign glands and the absence of basal cells is a strong indicator of acinar prostatic adenocarcinoma, the most common form of prostate cancer \cite{articleEpstein, trpkov2009usefulness}. Glandular tissue can objectively be selected as WOB or not WOB by using specific immunostaining for glandular tissue (Cytokeratin 8/18) and for basal cells (Cytokeratin 5/6 + p63) \cite{ma_method18}. We developed a data generation pipeline to semi automatically produce H\&E images with aligned WOB ground truth masks from the scanned immunofluorescence images. One important exception was made for intraductal cancer of the prostate (IDC-P). IDC-P is represented by high grade cancer (Gleason grade 4-5) inside a benign gland with basal cells. IDC-P cases were manually annotated as WOB. Then we used our DL Framework which reached the $2^{nd}$ best reported score in \textit{Camelyon17 Challenge} \cite{nicolas_camelyon18} to train models on such data.

In this paper, we demonstrate that DL models trained on such data are capable to predict potentially cancerous areas in biopsies with high accuracy. Finally, we compare the performance of the trained model against 17 biopsies annotated independently by three pathologists using only H\&E staining. The comparison demonstrate that the model performs as well as the pathologists. Even more, the model detects and accurately outlines existing WOB areas in two biopsies which were incorrectly annotated as totally WOB-free biopsies by three pathologists and in one biopsy by two pathologists.

\section{Related Work}

Recent advances in Artificial Intelligence (AI) and especially in DL have demonstrated high potential for automatic detection and classification of anomalies in Whole Slide Images (WSIs). For example, detection and classification of breast cancer \cite{breast_cancer_dl17, Liu_gigapixel17, nicolas_camelyon18}, lung cancer \cite{Coudray2018, nikolay_lung18}, colon cancer \cite{Mohammed2018YNetAD}, and prostate cancer \cite{prostate_biopsies18, Arvaniti2018, google_paper18} are the tasks where DL systems can help pathologists.

Several AI based systems for predicting anomalies in prostate tissues have been introduced recently. These systems include cancer detection in needle core biopsies \cite{prostate_biopsies18}, Gleason grading of tissue microarrays \cite{Arvaniti2018}, and Gleason grading in prostatectomies \cite{google_paper18}.

The authors in \cite{prostate_biopsies18} used multi-instance learning approach to train a classification network on a large dataset of 12,160 slides of digitalized needle biopsies. The chosen approach allowed to avoid the expensive pixel-wise annotations. However, it is hard to access the performance of the predictions on the pixel-level since the predictions were performed and evaluated on the slide-level. On contrary, in our paper, we produce and evaluate predictions for biopsies on the pixel-level.

The research conducted in \cite{Arvaniti2018} demonstrated that a DL model trained on Gleason graded tissue microarrays was capable to produce predictions comparable with the inter-pathologist agreement (kappa=0.71). However, the model was trained only on tissue microarrays which are outside of routine clinical workflow of the pathologist; the cores for tissue microarrays represent only small tumor regions and are used for research purposes only. On the other hand, we demonstrate the performance of our DL models on needle biopsies which are of explicit value to the pathologist's routine clinical workflow.

In \cite{google_paper18} the authors proposed using DL for classifying the Gleason score of WSIs. The diagnostic accuracy of 0.70 (p=0.002), higher than for pathologists, was reported. The authors also pointed out that the current Gleason can be refined to more finely characterize and describe tumor morphology quantitatively. The authors used a modified version of \textit{InceptionV3}  \cite{InceptionV3} to classify input image patches. Also, ensembling and hard example mining were employed to improve performance of trained models. In contrast, we formulate the problem as a binary segmentation task and use unet \cite{ronneberger2015} to produce pixel-level predictions. We provide pixels level segmentation results for biopsies while in \cite{google_paper18} the authors concentrated only on prostatectomies and slide-level predictions.

\section{The Deep Learning Framework}
\label{deep_learning_framework}

We developed a DL framework using TensorFlow \cite{tensorflow2015-whitepaper} for the training of Deep Neural Networks (DNNs) on Whole Slide Images (WSIs). The framework achieved the $2^{nd}$ best reported score in \textit{Camelyon17 Challenge} \cite{nicolas_camelyon18}. The key features of the framework are \textbf{quasi online hard example mining}, \textbf{compounding of semantic segmentation networks}, and extensive \textbf{data augmentation}.

\paragraph{Quasi Online Hard Example Mining.}
A WSI can contain an order of $10^9$ pixels, making it impossible to train DL networks directly on the full image. Thus, we trained our networks on smaller image patches. It was shown \cite{google_paper18, nicolas_camelyon18} that the patch sampling strategy during training has a great impact on the final performance of the model. We developed a training pipeline that enables \textit{quasi online hard example mining} sampling strategy. We trained our models with patches dynamically extracted from WSIs using a pixel-level probability density function inferred from the error of the model on the pixel-level classification.
%: $$\textit{error}_\theta(i) = 1-\textit{pred}_\theta(i)\cdot\textit{label}(i)$$
%where $\textit{pred}_\sigma(i)$ and $\textit{label}(i)$ are vectors representing respectively the model's predicted classes probabilities using parameters $\theta$ and the ground truth probabilities\footnote{usually an one hot vector} for the pixel $i$. The unnormalized log probability to sample a patch centered on the pixel $i$ is: $$E^\alpha_\theta(i)=\frac{\log(\alpha)}{\log(2)}\log(1+\textit{error}_\theta(i))$$
%The probability expression is then:
%$$P^\alpha_\theta(i) = \frac{e^{E^\alpha_\theta(i)}}{\sum_j e^{E^\alpha_\theta(j)}}$$ 
%where $\alpha$ is the maximum mass factor between the minimum and maximum errors. For instance, let's consider two pixels $i_m$ and $i_M$ that expose respectively a minimal and a maximal error. We have $\textit{error}_\theta(i_m)=0$, $\textit{error}_\theta(i_M)=1$ and  $\frac{P^\alpha_\theta(i_M)}{P^\alpha_\theta(i_m)} = \alpha$. For $\alpha=1$, all pixel probabilities are equal, i.e. $P^1_\theta$ is uniform.
The probability density function is computed on a separate process synchronously with the training process. This allows the hard example mining to be performed on frequent training cycles efficiently. See Figure \ref{fig:online_sampling} in Appendix for more details.

\paragraph{Semantic Segmentation Networks.} 
Semantic segmentation networks predict one of the class labels for every pixel of a patch. While our framework can handle a large variety of segmentation networks, in this study we used the state-of-the-art \textbf{unet}  \cite{ronneberger2015}. This model has an encoder for downsampling and a decoder for upsampling, which are linked with lateral skip connections. The network has high capacity of feature representation, thus, it can learn and aggregate knowledge at multiple scales in the data. Unet has become one of the most popular networks for semantic segmentation problems in biomedical imaging \cite{nikolay_lung18, Li2018, nicolas_camelyon18}.

\paragraph{Using Compound Model.} 
In order to increase the receptive field of the network, we developed a compound model approach which allows to expose the trained model to different resolution levels in WSIs. We combine two \textbf{unet} models learned from 1 and 2 mpp\footnote{micrometer per pixel} resolutions. Let $M^1$ be a \textbf{unet} model trained on 1 mpp. The network maps an RGB pixel $i \in [0,255]^3$ to the probability $M^1(i) \in [0,1]$ of belonging to the WOB class. Then let $M^{1,2}$ be another \textbf{unet} model trained on a resolution of 2 mpp using the output of $M^1$ as an extra input channel as illustrated in Figure \ref{fig:compound_model}. Compound models described above bring the benefits of ensemble learning: we can learn the individual \textbf{unet} models with different strategies or hyper-parameters inducing different expressiveness of models that can eventually be integrated as their combination in the compound model.

\paragraph{Data Augmentation.} 
The difficulty of producing large amount of annotated data can be leveraged with an extensive usage of data augmentation. Our framework allows to program a data processing pipeline in a simple text based configuration file. Different transformation operators can be piped to produce a final augmented patch. The augmentation is performed online during the training. The augmentation operators include rotation, mirroring, elastic deformation, color jittering. Table \ref{tab:augmentations} in Appendix provides augmentation examples.

\section{Data Generation}
In order to obtain a more objective ground truth, we introduced the concept of WOB, `glandular tissue WithOut Basal cells'. The objectivity of WOB is in the fact that the presence of basal cells can be assessed by using immunohistochemical markers. Presence of basal cells indicates that a gland is healthy which implies that WOB corresponds to potentially malignant or cancerous structures \cite{articleEpstein, trpkov2009usefulness}. For our experiments, we produced three WOB annotated datasets described below (more details on the datasets are in Table \ref{tab:data_summary}).

\subsection{Semi-automatically WOB annotated prostatectomies} 
WOB areas can be accurately segmented out in the prostate tissues using a method described in \cite{ma_method_patent18, ma_method18}. The method is based on staining prostate tissues towards basal cells with Cytokeratin 5/6, then epithelial tissue with Cytokeratin 8/18, cancerous cells with Alpha-methylacyl-CoA racemase (AMACR), and nuclei with 4',6-diamidino-2-phenylindole (DAPI) \cite{ma_method18}. Then the same slides are stained with H\&E and scanned at a high resolution. The immunofluorescent stainings mark specific structures in the prostate tissues which can be automatically converted into the WOB areas. By overlaying the stainings with H\&E, the WOB areas exactly match the corresponding WOB areas in an H\&E image which allows us to generate accurate and detailed WOB annotations.

\paragraph{Converting immunofluorescence images into annotations.} 
We developed an algorithm to generate binary ground truth masks from scanned immunofluorescence images. For the different immunofluorescence channels, the information are not present in the entire gland areas but localized to certain positions within the gland. As a result, the markers of the different immunochannels consequently appear at different places within the gland. To make an overall segmentation of the entire gland area, this local information is propagated over the gland by the density estimation filter which allowed to distribute the local information evenly within a local neighborhood of the image. The density estimation filter was applied to all three immunofluorescence channels (Cytokeratin 5/6, Cytokeratin 8/18, and AMACR). Then two heatmaps were generated. The first one mapped epithelial versus epithelial plus basal channel. The second mapped AMACR versus epithelial plus AMACR channel. Both heatmaps produced an estimate in the range $[0,1]$ where high intensity indicated high probability for the WOB class. The design of the heatmaps is based on the fact that adenocarcionoma only occurs within epithelial tissue and the relation between the epithelial channel to the basal cell and AMACR channels consequently indicate the WOB or not WOB class. Finally, these two heatmaps were combined to generate a best possible WOB mask (see Figure \ref{example_ma_process} for a WOB mask example). To our experience the AMACR heatmap was less reliable compared to the basal cell heatmap and when they were inconsistent the basal cell heatmap was used.

\paragraph{Reviewing the automatically generated annotations.} 
We decided to ask pathologists to review the automatically generated masks. The main reason was that the intraductal cancer of the prostate (IDC-P) represented by high grade cancer (Gleason grade 4-5) inside a benign gland with basal cells must be manually annotated. The reviews led to some minimal manual corrections in the automatically generated masks. In total, $48$ prostatectomy masks were generated and reviewed.

\paragraph{Using consecutive slices.} 
We included $48$ consecutive H\&E stainings to the original $48$ H\&E prostatectomies in order to further increase the size of trained data and eventually the performance of trained models. A few consecutive slices were excluded since the difference in morphology was not acceptably negligible. The ground truth for the consecutive images was obtained by registering the corresponding H\&E image and then applying the calculated transformation to the ground truth of the original H\&E images (see Figure \ref{fig:registrations_examples} for an example). The registration was done using non rigid registration with elastix software (http://elastix.isi.uu.nl/).

\paragraph{Extending scanners variation.} 
All H\&E stainings were scanned with three commercially available scanners. The reason for scanning the images by the three scanners was the fact that we wanted the model to be robust across these scanners. The scanned images were registered to the original images and a corresponding ground truths were produced.

\paragraph{Final dataset.} We produced $295$ images exhaustively annotated with WOB; the images represented variations in prostate cancer stages, scanners, and tissue morphology. This dataset was used for training DL models.

\subsection{Manually annotated WOB biopsies supported by H\&E and DAB stainings} 
The second dataset consisted of 181 biopsies scanned with two different scanners at 0.5 and 0.22 micrometer per pixel accordingly. All biopsies were exhaustively annotated by six pathologists. All the pathologists who annotated the images were carefully selected. The criteria was that the pathologists had proper specialist training in pathology as well as several years of clinical practice in pathology. The number of years of experience varied between 5-20 for both groups. The pathologists used AMACR, a staining for cancerous cells, and p63, a staining for basal cells. This dataset was split into train set of 118 biopsies and 63 biopsies for test. The train set was used for training and finetuning DL models and the test set was used for evaluation of the trained models.

\subsection{Manually annotated WOB biopsies supported by H\&E staining only} 
Finally, 17 out of 181 annotated biopsies, 13 with WOB and 4 without WOB, were exhaustively annotated by other three pathologists with H\&E staining only, without any support of p63 and AMACR stainings. This dataset was used for comparison of DL models to three pathologists.

\section{Experimental Setup and Results}
\label{experimental_conditions}
In total, we trained and evaluated six DL models on a cluster with 10 TitanXp GPUs with 12GB VRAM and 64GB RAM. Training all the six models required approximately five days.

\subsection{Trained DL models}
We trained DL models (1) only on prostatectomies, (2) only on biopsies, and (3) on prostatectomies first and then finetuned on biopsies. For each dataset we trained two models, a model $M^1$ on 1 mpp and a \textit{compound model} $M^{1,2}$ on 1 mpp and 2 mpp (see Figure \ref{fig:compound_model}). 

The model $M^1_{pr}$ was trained only on $1$ mpp prostatectomies, $M^{1,2}_{pr}$ was trained on $1$ and then $2$ mpp prostatectomies; the model $M^1_{bi}$ and $M^{1,2}_{bi}$ were trained only on biospies on $1$ mpp and on both $1$ mpp and $2$ mpp correspondingly. Finally, $M^1_{pr,bi}$ and $M^{1,2}_{pr,bi}$ were trained on prostatectomies first and then finetuned on biopsies at $1$ mpp and then both $1$ and $2$ mpp.

Each model was trained for $10^{6}$ iterations. Each iteration had a batch of 32 patches. The patches were sampled using \textit{quasi online hard example mining} described in Appendix \ref{fig:online_sampling}. Each patch was 188 by 188 pixels. The relatively small size of the patches constrained the sampling areas of hard regions without overshooting these.

\subsection{Evaluation setup}
The test set for the evaluation consisted of $63$ needle biopsies annotated with WOB by several pathologists with support of DAB stainings. We calculated precision-recall curves, areas under the precision-recall curve (PR AUC), and  maximum F1 scores for each model. The results are in Figure \ref{fig:results_all}. 

\begin{figure}
  \centering
  \includegraphics[width=0.33\textwidth,scale=1]{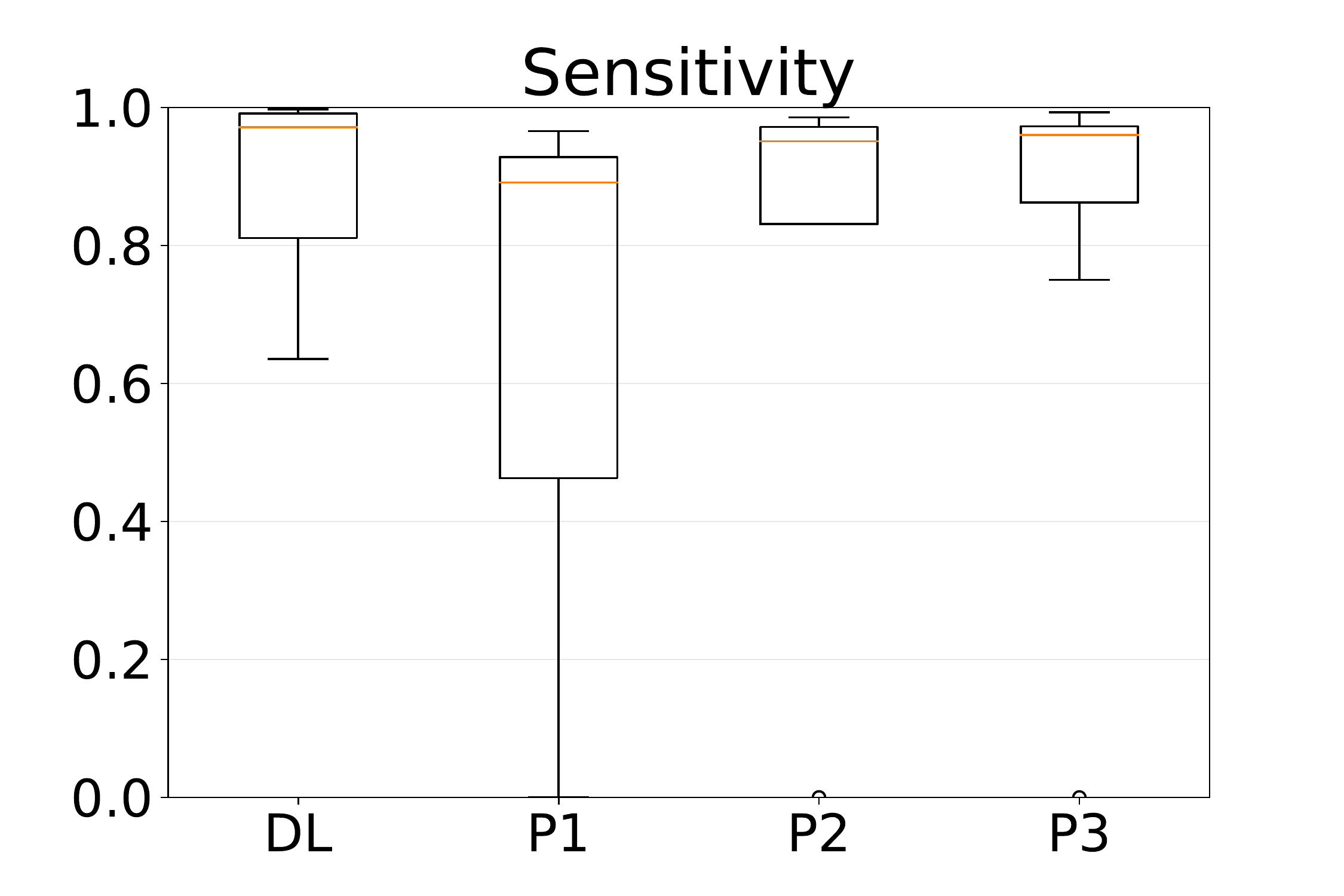}
  \hspace{-0.7cm}
  \includegraphics[width=0.33\textwidth,scale=1]{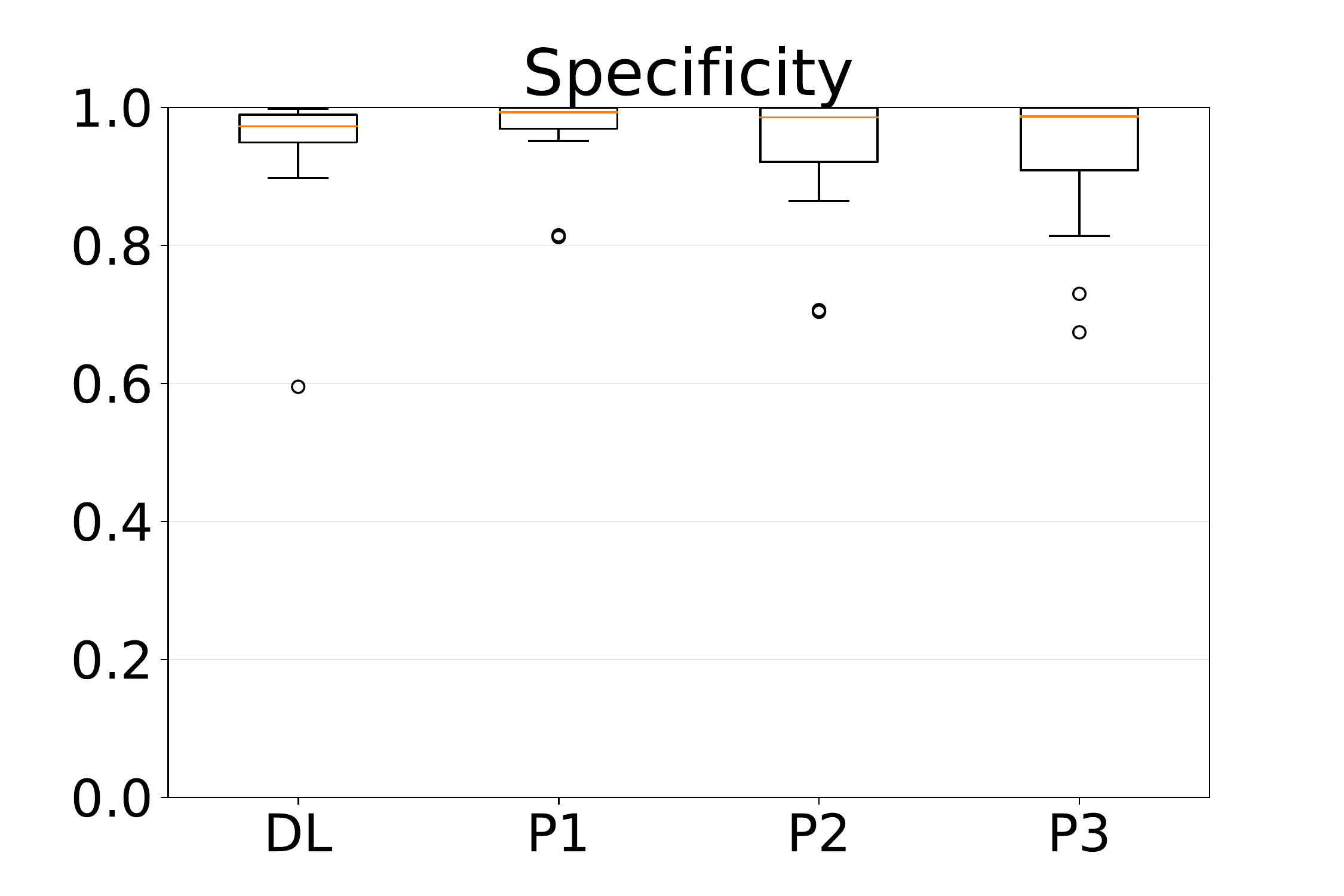}
  \hspace{-0.7cm}
  \includegraphics[width=0.33\textwidth,scale=1]{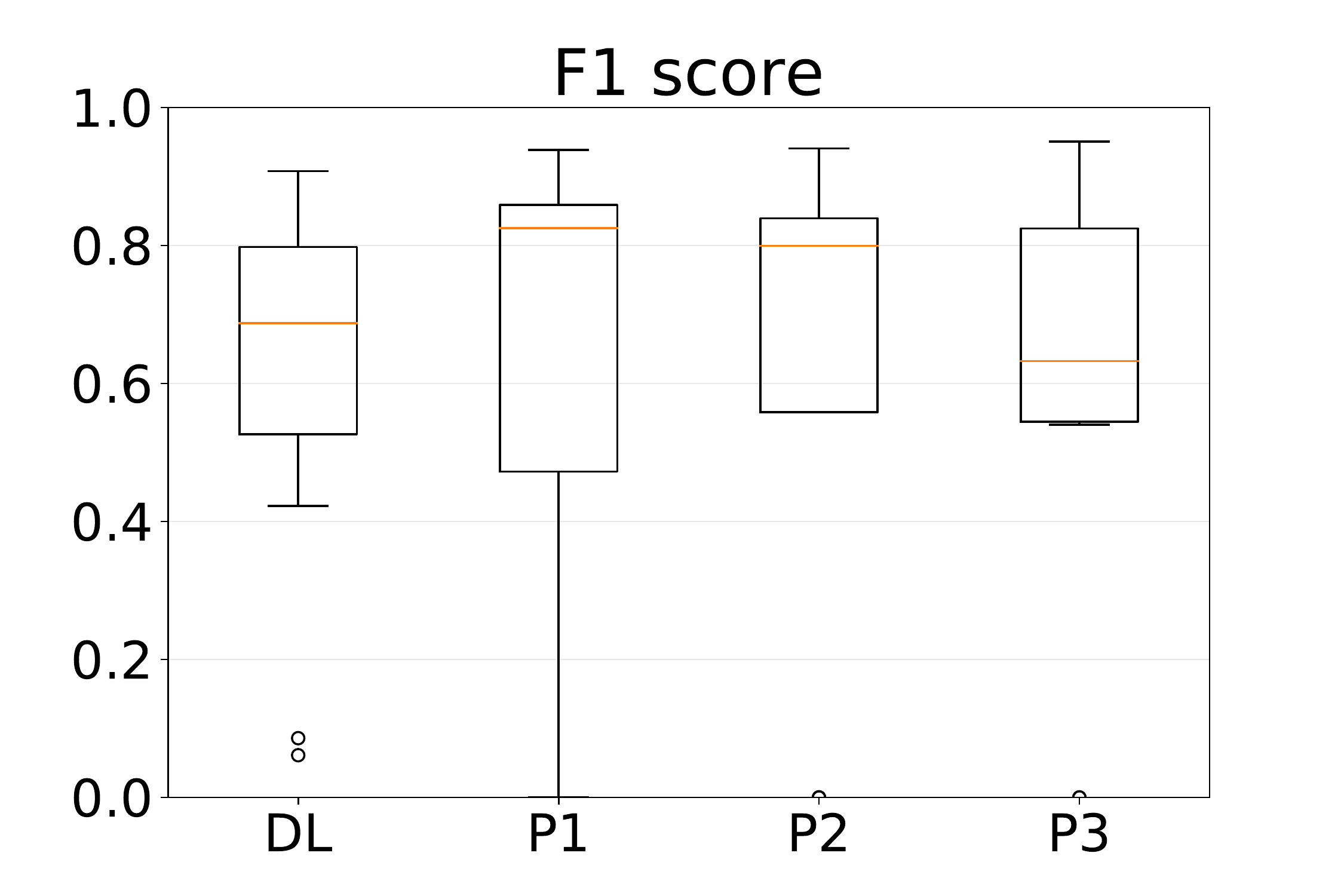}
  \caption{\textit{Sensitivity, specificity, and F1 score} for WOB predicted by the model $M^{1,2}_{pr,bi}$ (DL) and annotated by the three pathologists (P1, P2, P3) for $17$ biopsies. Sensitivity and F1 score were calculated only for $13$ WOB positive biopsies while specificity for all 17 biopsies.}
  \label{fig:results_biopsies17}
\end{figure}

Then we chose the best performing DL model, the compound model $M^{1,2}_{pr,bi}$, and compared its performance to three pathologists who manually annotated WOB areas in 17 biopsies using H\&E stainings only. The results are summarized in Figure \ref{fig:results_biopsies17}.

\subsection{Performance of DL models} 
\begin{figure}
  \centering
  \includegraphics[width=0.49\textwidth,scale=0.8]{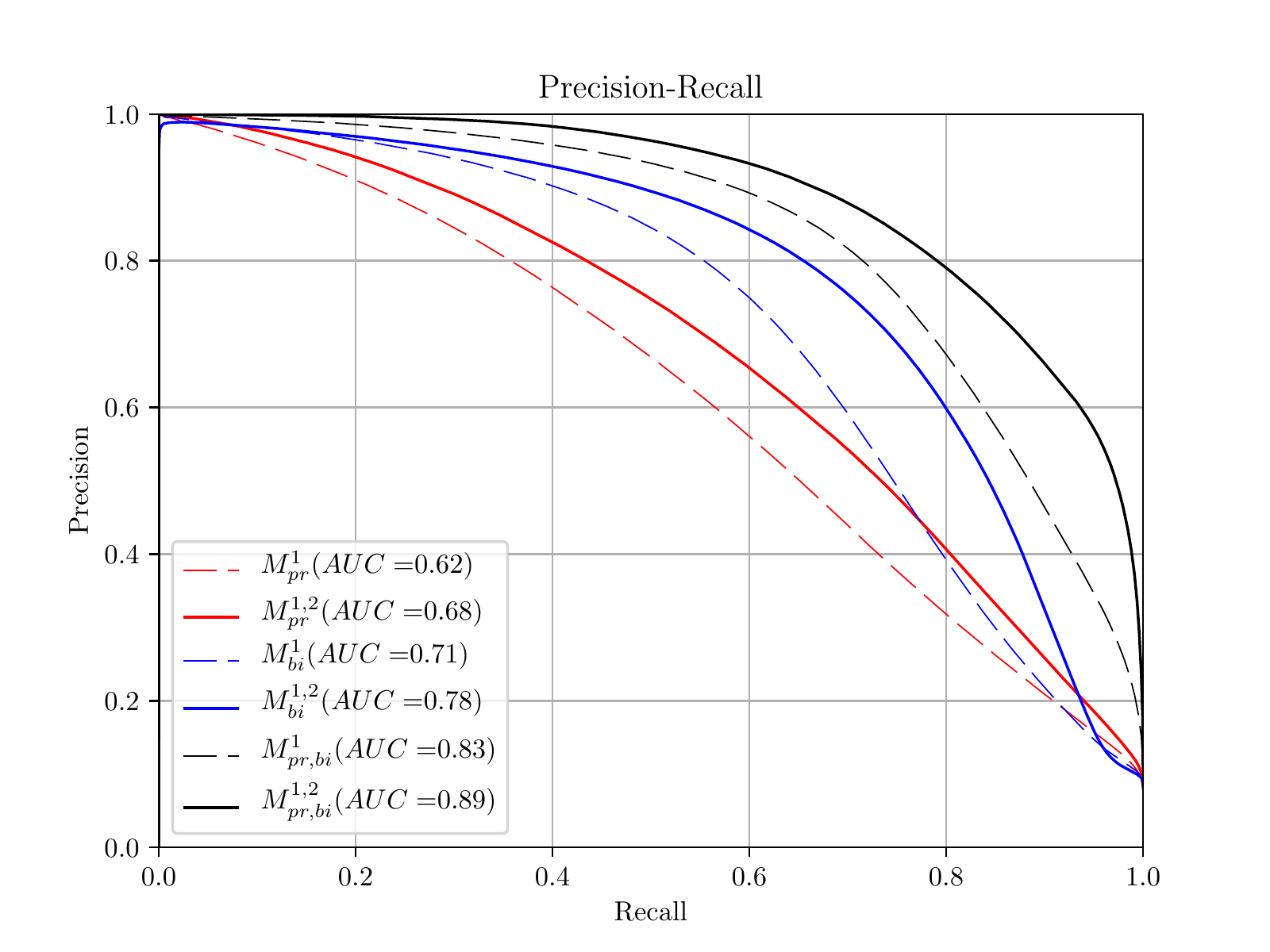}
  \includegraphics[width=0.49\textwidth,scale=0.8]{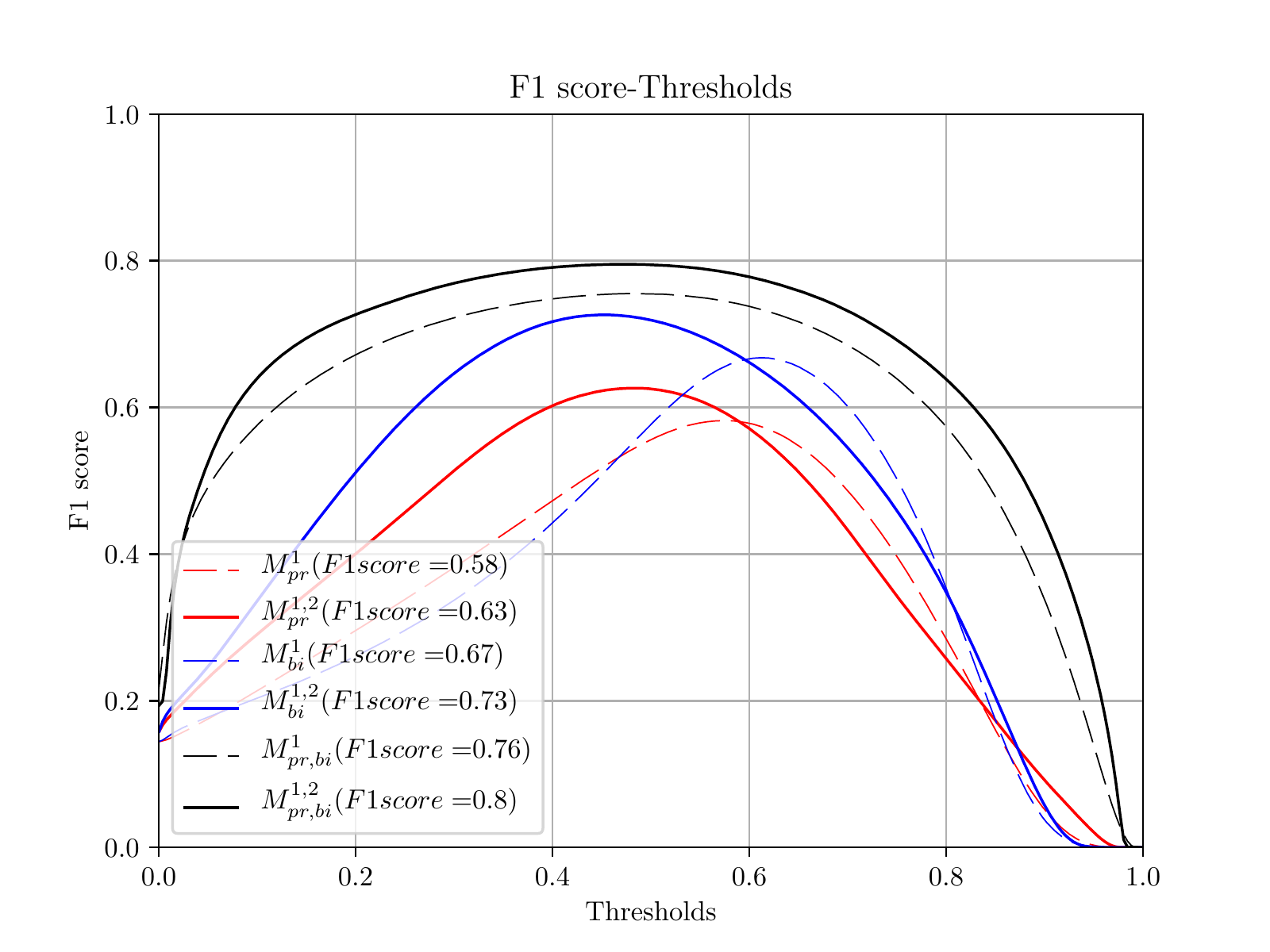}
  \caption{\textit{Precision-Recall (PR) curves} (left) and \textit{F1 score curves} (right) for all models. In the left plot, AUC stands for Area Under PR Curve (AUC) and is calculated for each model. In the right plot, a maximum achievable F1 score for each model is shown. $M^1_{pr}$ was trained only on $1$ mpp prostatectomies, $M^{1,2}_{pr}$ was a compound model trained on $1$ and then $2$ mpp prostatectomies; the model $M^1_{bi}$ and $M^{1,2}_{bi}$ were trained only on biopsies on $1$ mpp and on both $1$ mpp and $2$ mpp correspondingly. Finally, $M^1_{pr,bi}$ and $M^{1,2}_{pr,bi}$ were trained on prostatectomies first and then finetuned on biopsies at $1$ mpp and then both $1$ and $2$ mpp.}
  \label{fig:results_all}
\end{figure}

The predictions demonstrated high performance at pixel-level reaching F1 score of 0.80 and PR AUC of 0.89 for the model $M^{1,2}_{pr,bi}$ trained on prostatectomies and then finetuned on biopsies (see Figure \ref{fig:example_biopsy_prediction} for prediction examples by different models). We noticed that training compound models on $2$ mpp consequently after training models on $1$ mpp helped to remove false positives and to improve the performance of the models. The performance of the models $M^{1}_{pr}$ and $M^{1,2}_{pr}$ trained only on prostatectomies showed the worst results with PR AUCs of 0.62 and 0.68 accordingly (see Figure \ref{fig:results_all}, the left plot). The models $M^{1}_{bi}$ and $M^{1,2}_{bi}$ trained only on biopsies achieved higher PR AUCs of 0.71 and 0.78. Finally, the models $M^{1}_{pr,bi}$ and $M^{1,2}_{pr,bi}$ outperformed both biopsies and prostatectomies with PR AUCs of 0.83 and 0.89. The same order holds for F1 scores across the models. Also, the models trained on prostatectomies first and then finetuned on biopsies showed the widest F1 scores (see Figure \ref{fig:results_all}, the right plot) which implies that the models $M^{1}_{pr,bi}$ and $M^{1,2}_{pr,bi}$ are more robust to different thresholds.

\subsection{Comparison of the best performing DL model to three pathologists} 
Finally, we compared the predictions by the best performing model, the compound model $M^{1,2}_{pr,bi}$, with three pathologists who were instructed about the WOB class and then annotated WOB areas using only H\&E stainings. Remarkably, in two biopsies all three pathologists totally missed WOB areas but our model successfully detected those areas (see Figure \ref{fig:missed_wobs} in Appendix for examples of the missed areas). Two out of three pathologists missed all WOB areas in a third biopsy. Quantitatively, pathologists performed as well as the trained model (see Figure \ref{fig:results_biopsies17} for sensitivity, specificity, and F1 score boxplots). 

\section{Conclusions and Future Work}
\label{conclusions_and_future_work}

In this paper, we introduced a novel DL framework for segmenting potentially cancerous areas in prostate biopsies using semi-automatically generated data. Introducing WOB class allowed us to generate data with minimal involvement of pathologists for pixel-level annotations.

The performance of the trained models was tested on 63 biopsies from several clinical labs, scanners, and annotated by different pathologists. The trained models showed high potential in identifying potentially cancerous areas. We demonstrated that training the proposed compound models is beneficial for increasing the receptive field of the networks and eventually for the performance of the models. We showed that using semi-automatically generated data leads to increase in performance compared to models, trained only on biopsies.

We evaluated the best performing model against three pathologists who annotated 17 biopsies using H\&E staining only. The comparison demonstrated that the model performed on a par with the pathologists. Even more, the model detected and outlined existing WOB areas in three biopsies which were annotated as WOB-free biopsies by the three pathologists.

For future work, we plan to evaluate the performance of the trained models at a few clinical pathology labs. We are planing to collect feedback from pathologists who will use our product with the developed DL algorithm. Also, we are going to test the trained models on more diverse WSIs images stained and scanned in different labs in order to evaluate the robustness of the algorithm further. Finally, we are planning to build a mapping from WOB predictions to Gleason score grading scheme.

\section{Authors' Contributions}
N.P. developed the DL framework. N.B. and N.P. designed, carried out the experiments, and wrote the manuscript; N.B. evaluated the results. F.G. helped with annotation generation. M.A. developed the method to generate the annotations. M.A. and D.H. contributed to data augmentation and experiment design. L.B. and K.E. introduced WOB concept, designed and organized data generation process. M.H., L.K., and C.S. supervised the project.

\midlacknowledgments{We would like to thank pathologists for providing the data and expertise to the project. We thank David Buffoni for the comments on the manuscript.}
\bibliography{burlutskiy19} % Entries are in the "refs.bib" file

\appendix
\renewcommand\thefigure{\thesection.\arabic{figure}}    
\renewcommand\thetable{\thesection.\arabic{table}}    
\setcounter{figure}{0} 
\setcounter{table}{0} 

\section{Supplementary Material}

\begin{figure}[H]
\centering
\begin{tabular}{cc}

  \includegraphics[scale=0.9]{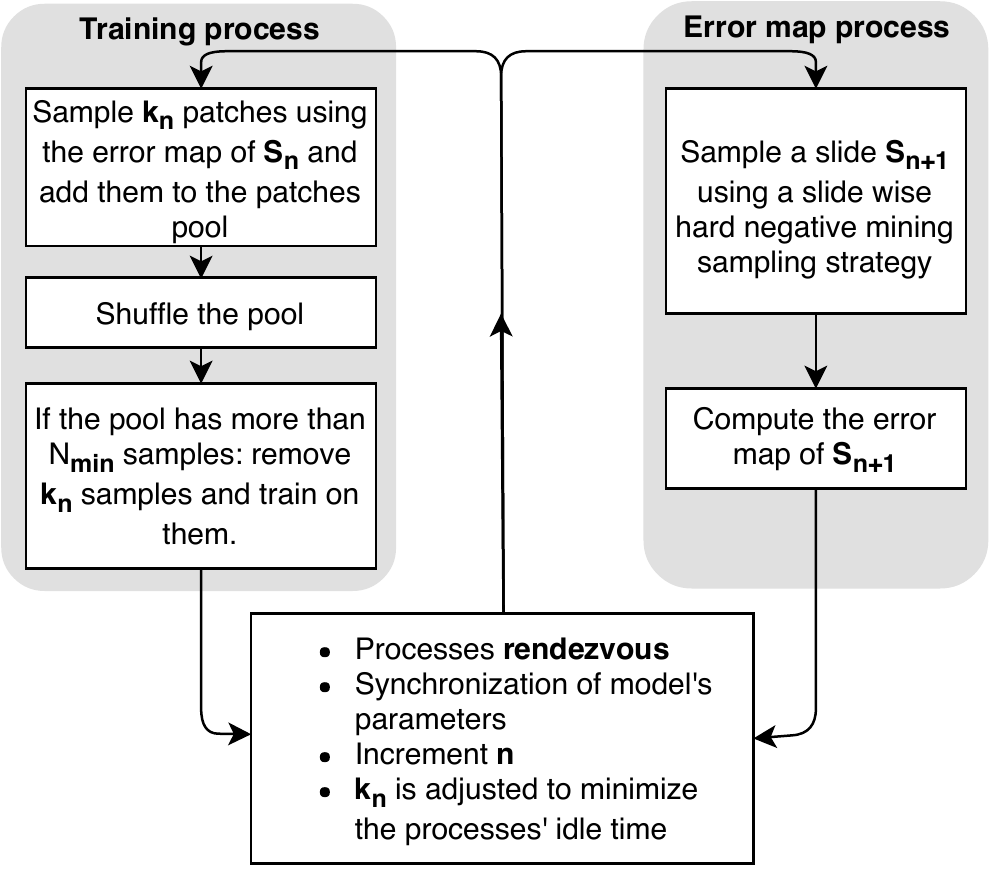} &
  \begin{tabular}[b]{c}
    Annotation \\
  \includegraphics[scale=0.45]{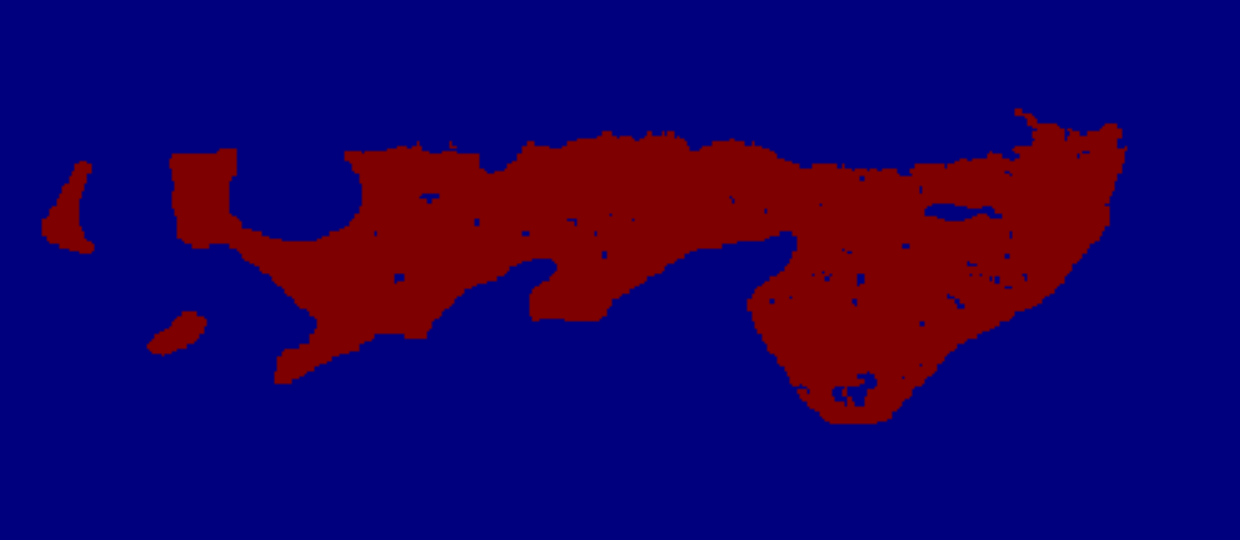} \\

  Prediction \\
  \includegraphics[scale=0.45]{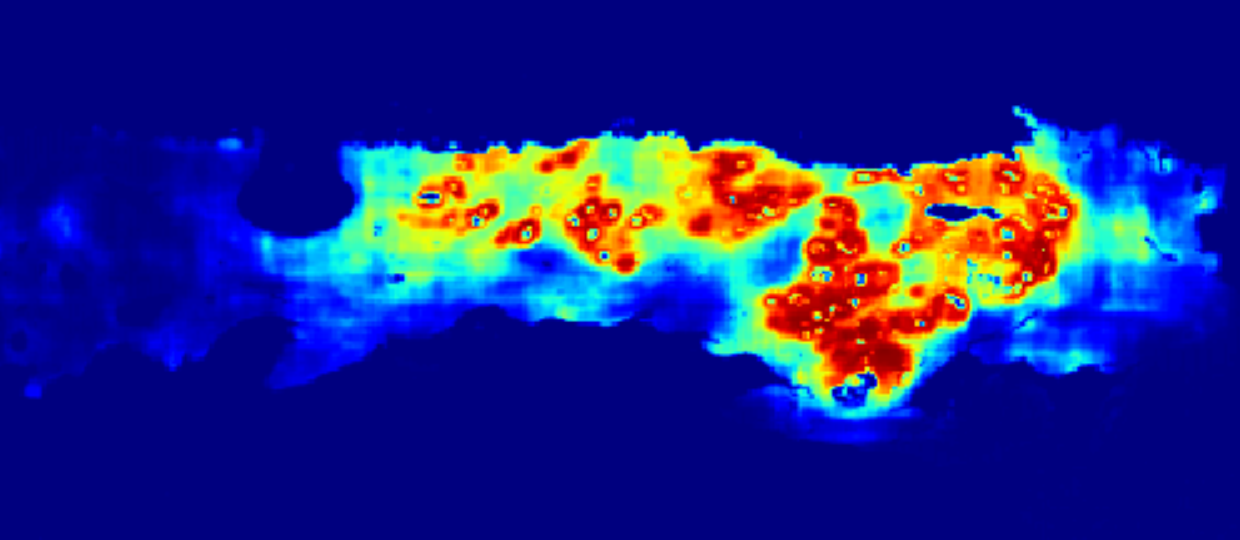} \\

  Error map\\
    \includegraphics[scale=0.45]{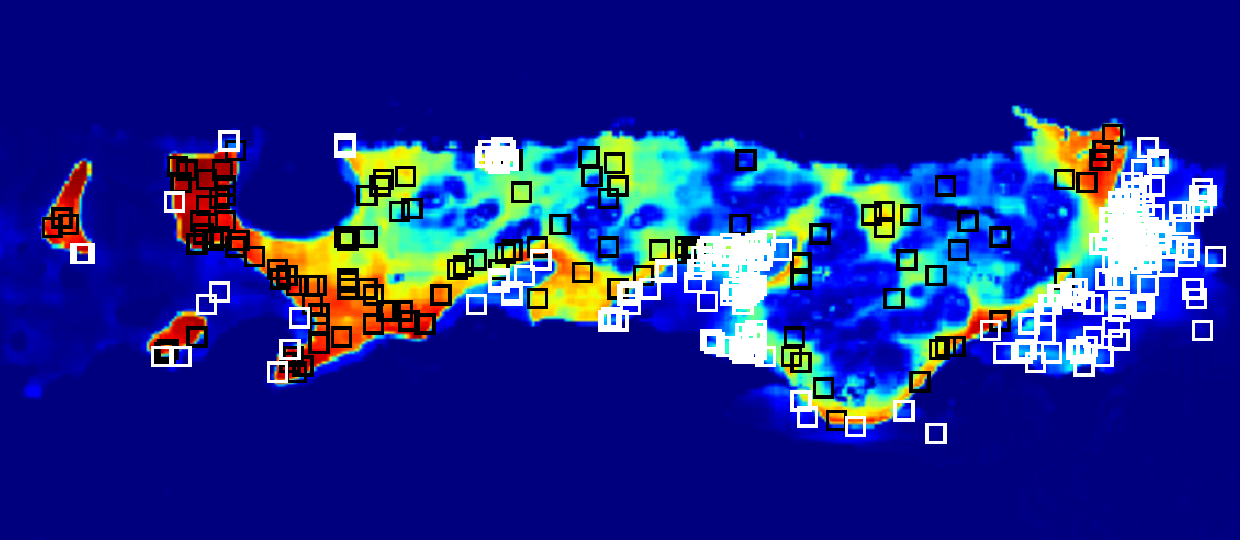} \\

  \end{tabular} \\
  (a) & (b)\\

\end{tabular}
\caption{(a) A flowchart explaining the training with quasi online hard example mining of patches. The \textbf{error map process} starts by computing the error map of an initial $S_0$ slide. Then the \textbf{training process} starts and uses the computed error map to sample $k_0$ patches from $S_0$ that are fed into a pool. The two processes rendezvous and the cycle continues for $S_1,S_2,...$ until the pool reaches a minimal size $N_{min}$ to allow the building of mini-batches containing uncorrelated patches. Then the \textbf{training process} starts to update the model parameters by training on $k_n$ patches sampled from the pool. Once the \textbf{training process} meets the \textbf{error map process} at the rendezvous, the model parameters are synchronized. Then a new cycle begins, the \textbf{error map process} samples $S_{n+1}$ and computes its error map with the updated parameters while the \textbf{training process} samples patches from $S_n$. The \textbf{error map process} samples $S_{n+1}$ using a slide wise hard example mining strategy, e.g. a slide among those containing many false positives is sampled. At each rendezvous, $k_n$ is the adjusted version of $k_{n-1}$ to minimize the idle time. Depending on which process is idling, $k_n$ is increased for the \textbf{training process} or decreased for the \textbf{error map process}. The two processes can run on different hardware devices and only need to communicate over the network. (b) An example of a corresponding annotation mask, prediction heat map and error map for a biopsy area. The error map is overlaid with examples of randomly sampled patches for \textit{not WOB} (white) and \textit{WOB} (black) classes. The patches are more likely to be sampled from high error regions.
}
\label{fig:online_sampling}
\end{figure}

\begin{table}
\centering
\caption{Augmentations applied on a patch. An overlaid grid illustrates the deformations.}
\begin{tabular}{|l|l|l|l|}
\hline Rotation and mirroring & Elastic & Color & All \\  
	\includegraphics[width=0.24\columnwidth]{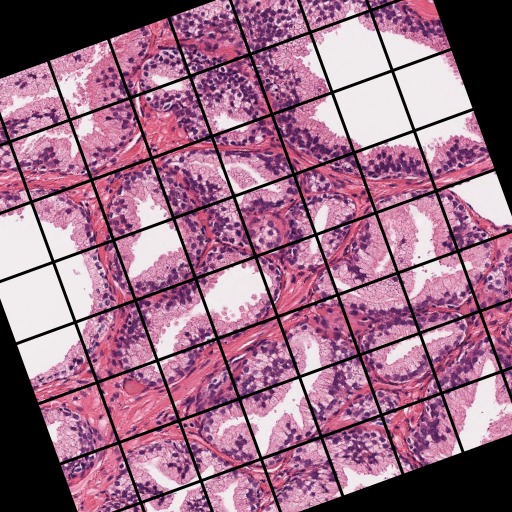} &
	\includegraphics[width=0.24\columnwidth]{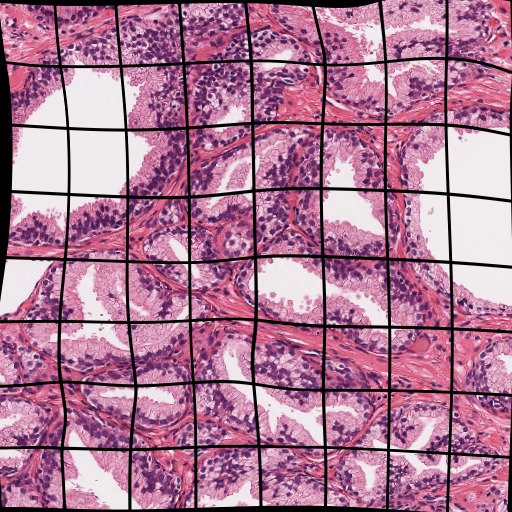} &
	\includegraphics[width=0.24\columnwidth]{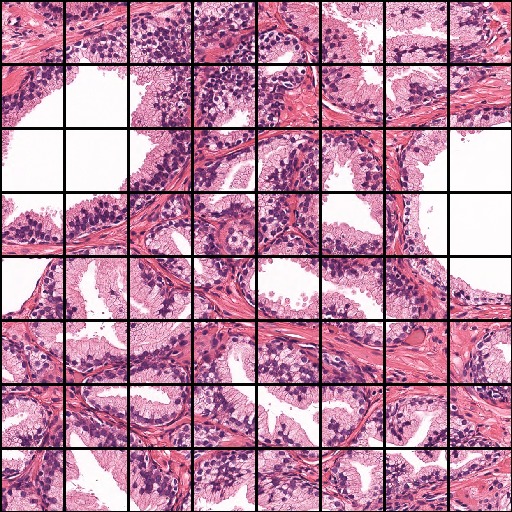} &
	\includegraphics[width=0.24\columnwidth]{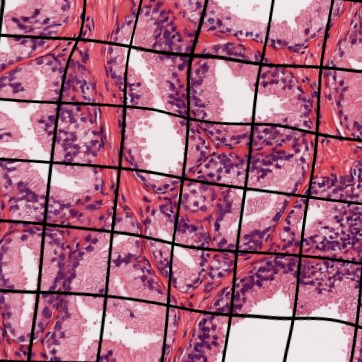}
	\\ 
	
	\includegraphics[width=0.24\columnwidth]{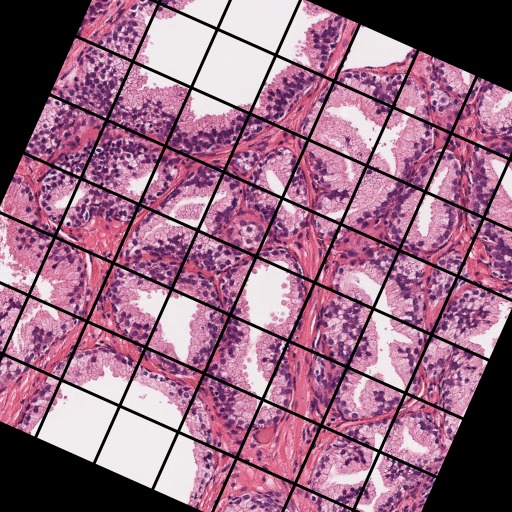} &
	\includegraphics[width=0.24\columnwidth]{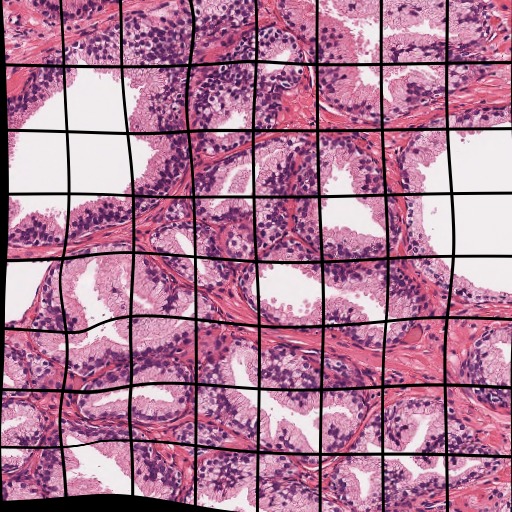} &
	\includegraphics[width=0.24\columnwidth]{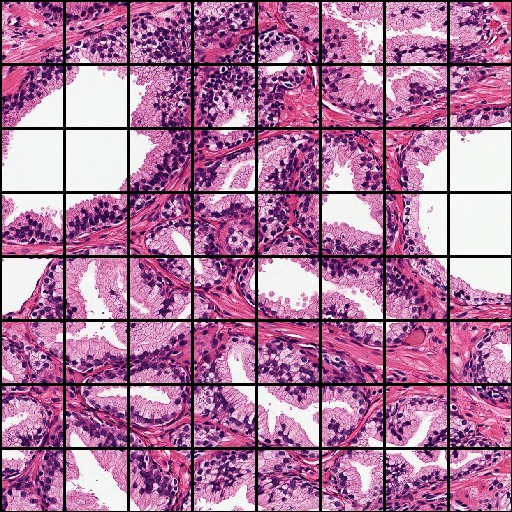} &
	\includegraphics[width=0.24\columnwidth]{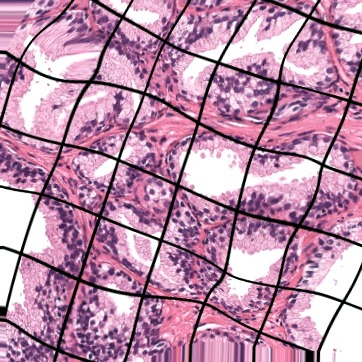}
	\\ 
	
	\includegraphics[width=0.24\columnwidth]{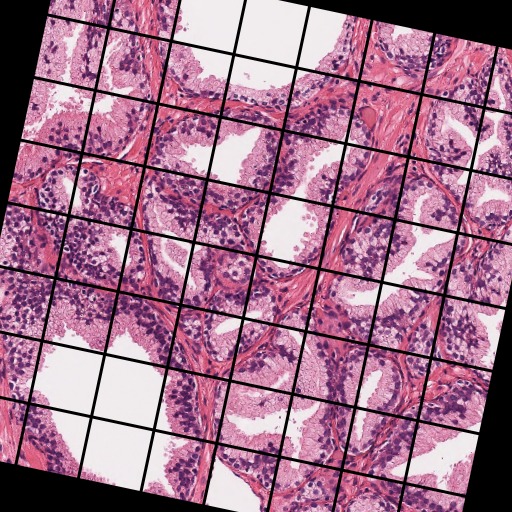} &
	\includegraphics[width=0.24\columnwidth]{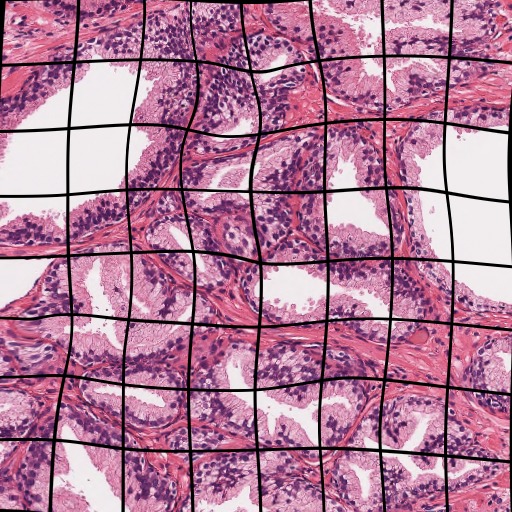} &
	\includegraphics[width=0.24\columnwidth]{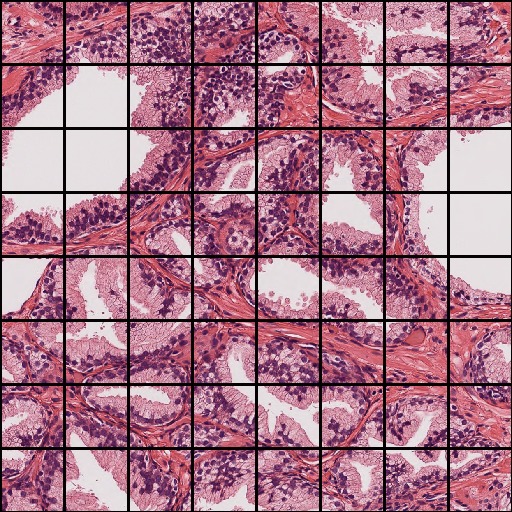} &
	\includegraphics[width=0.24\columnwidth]{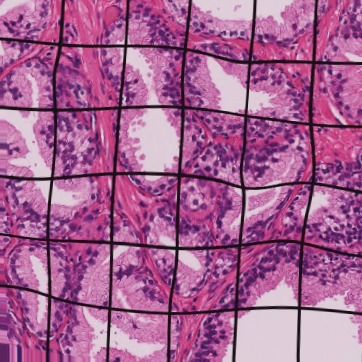}
	\\ 
	
	\includegraphics[width=0.24\columnwidth]{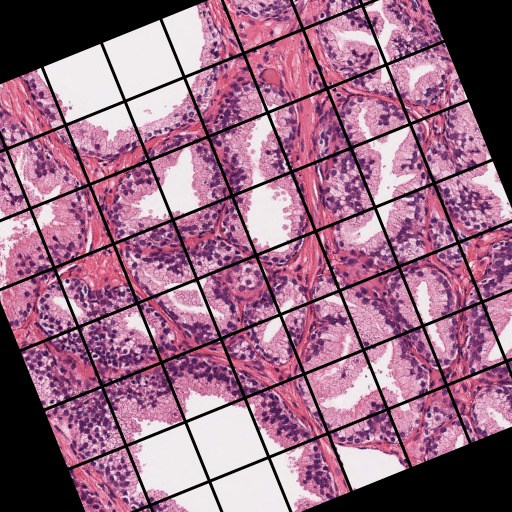} &
	\includegraphics[width=0.24\columnwidth]{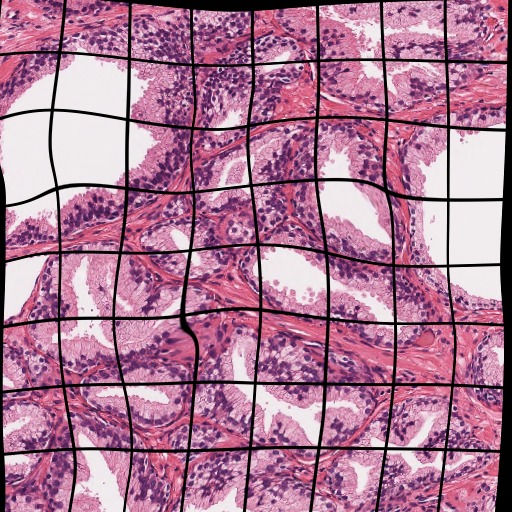} &
	\includegraphics[width=0.24\columnwidth]{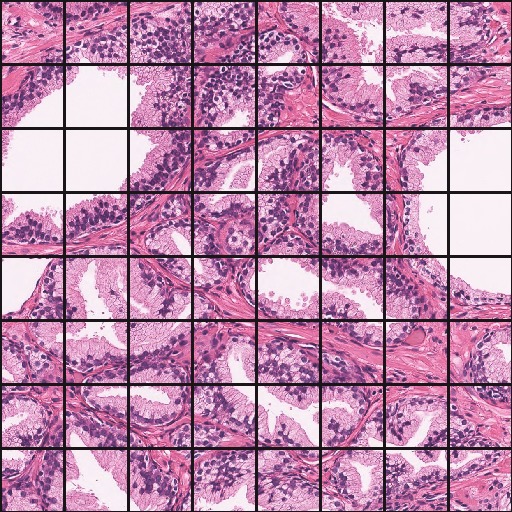} &
	\includegraphics[width=0.24\columnwidth]{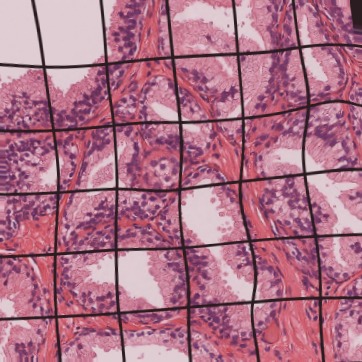}
	\\ 
	
	\includegraphics[width=0.24\columnwidth]{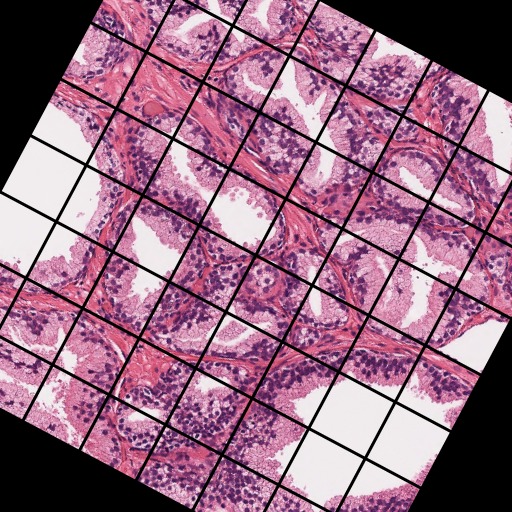} &
	\includegraphics[width=0.24\columnwidth]{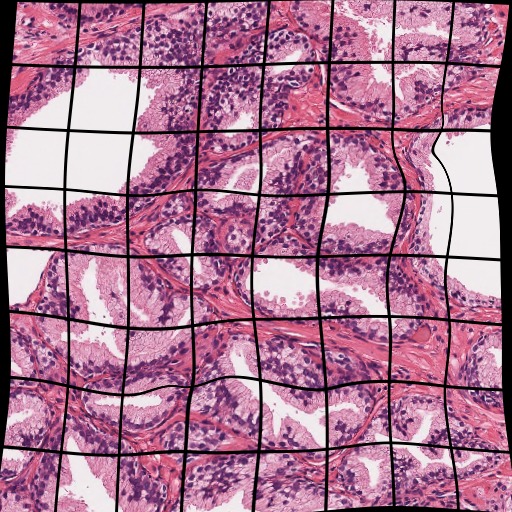} &
	\includegraphics[width=0.24\columnwidth]{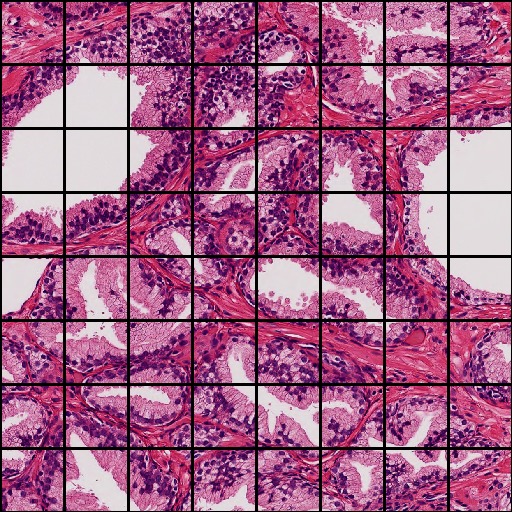} &
	\includegraphics[width=0.24\columnwidth]{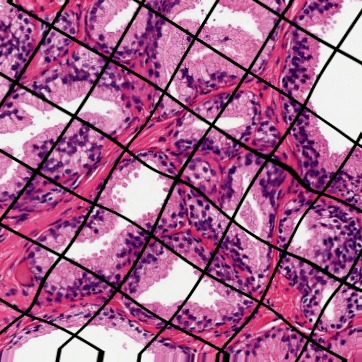}
	\\ \hline
\end{tabular}
\label{tab:augmentations}
\end{table}

\begin{figure}[H]
  \centering
  \includegraphics[width=0.49\textwidth,scale=0.8]{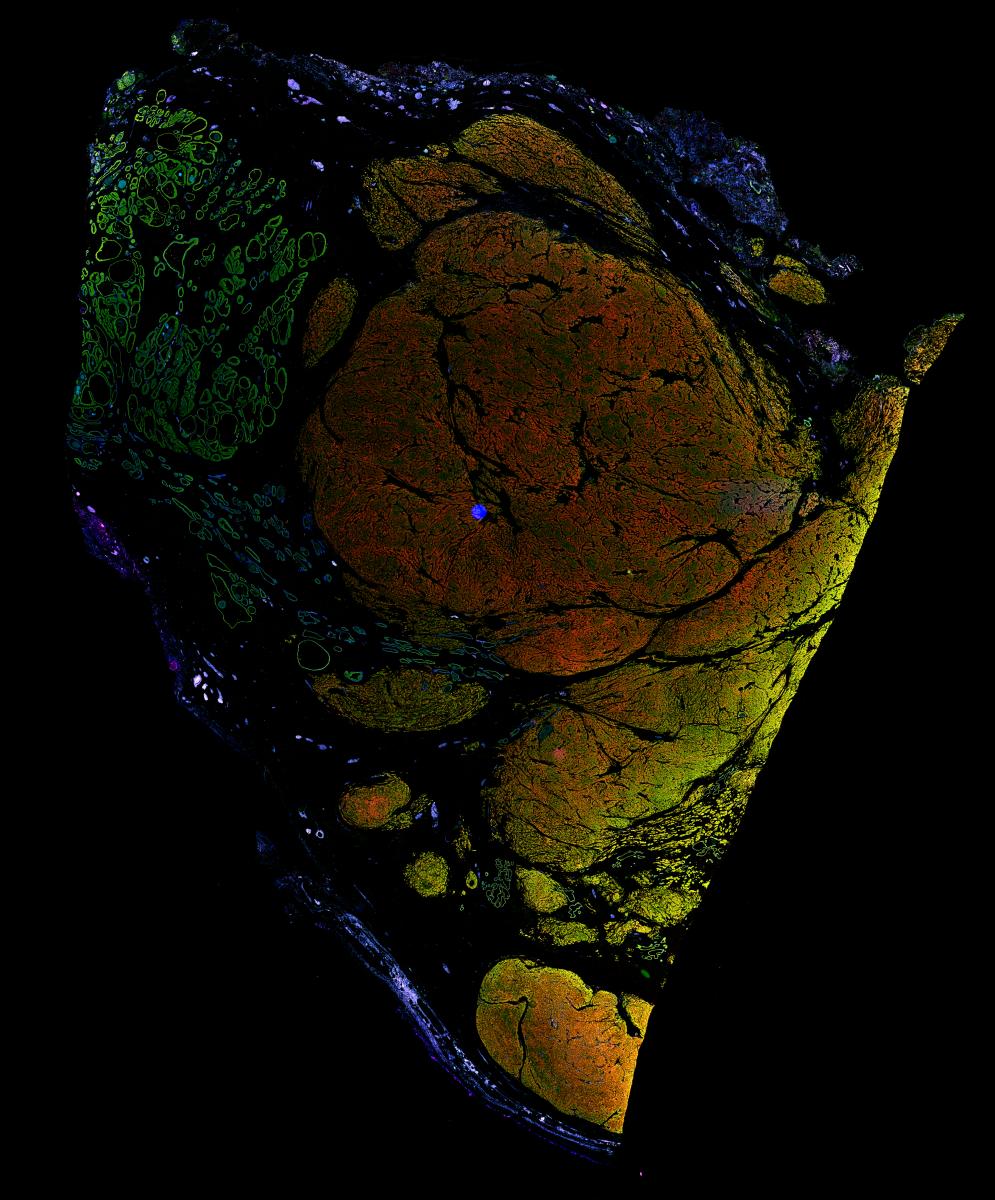}
  \includegraphics[width=0.49\textwidth,scale=0.8]{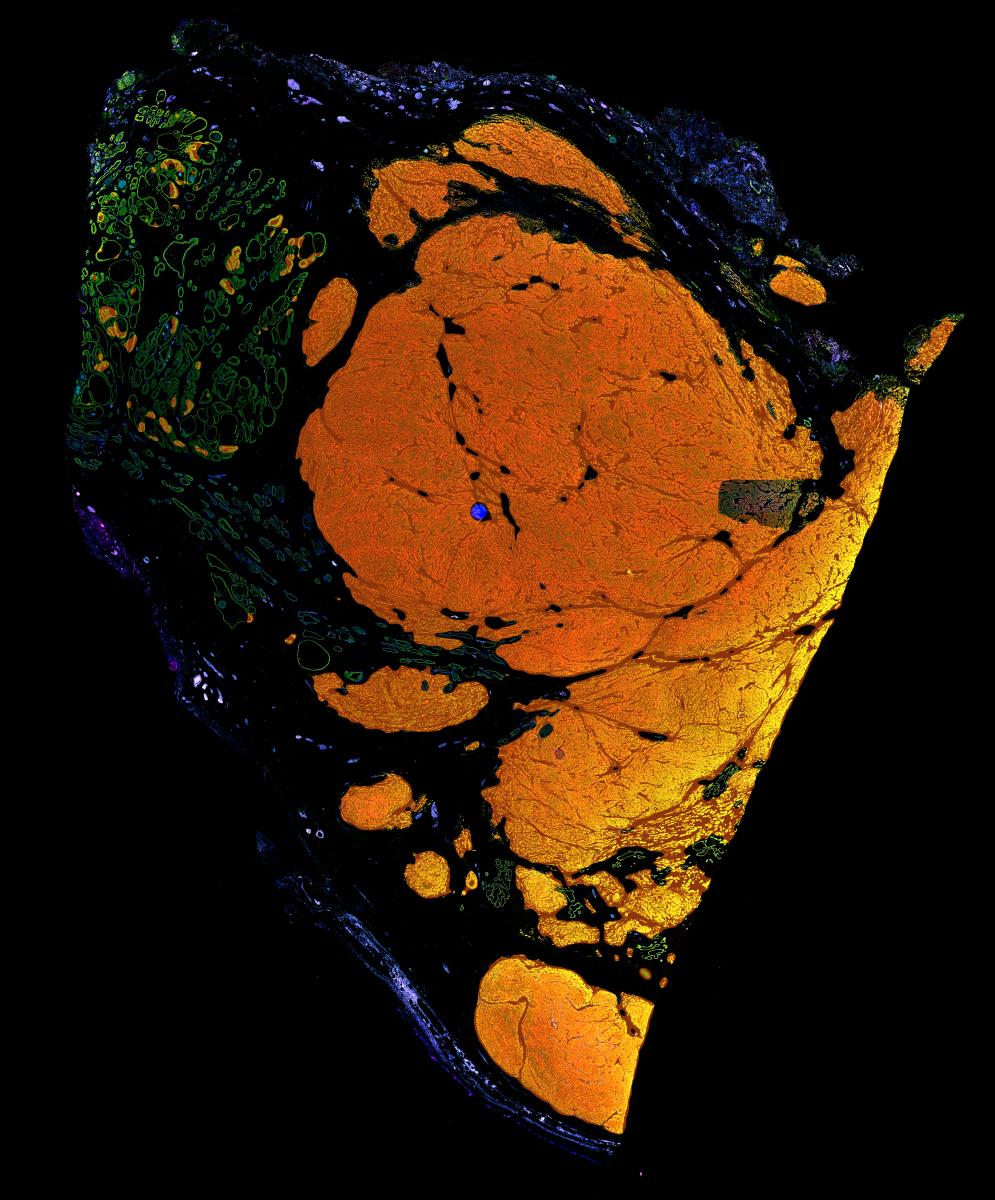}
    \includegraphics[width=0.49\textwidth,scale=0.8]{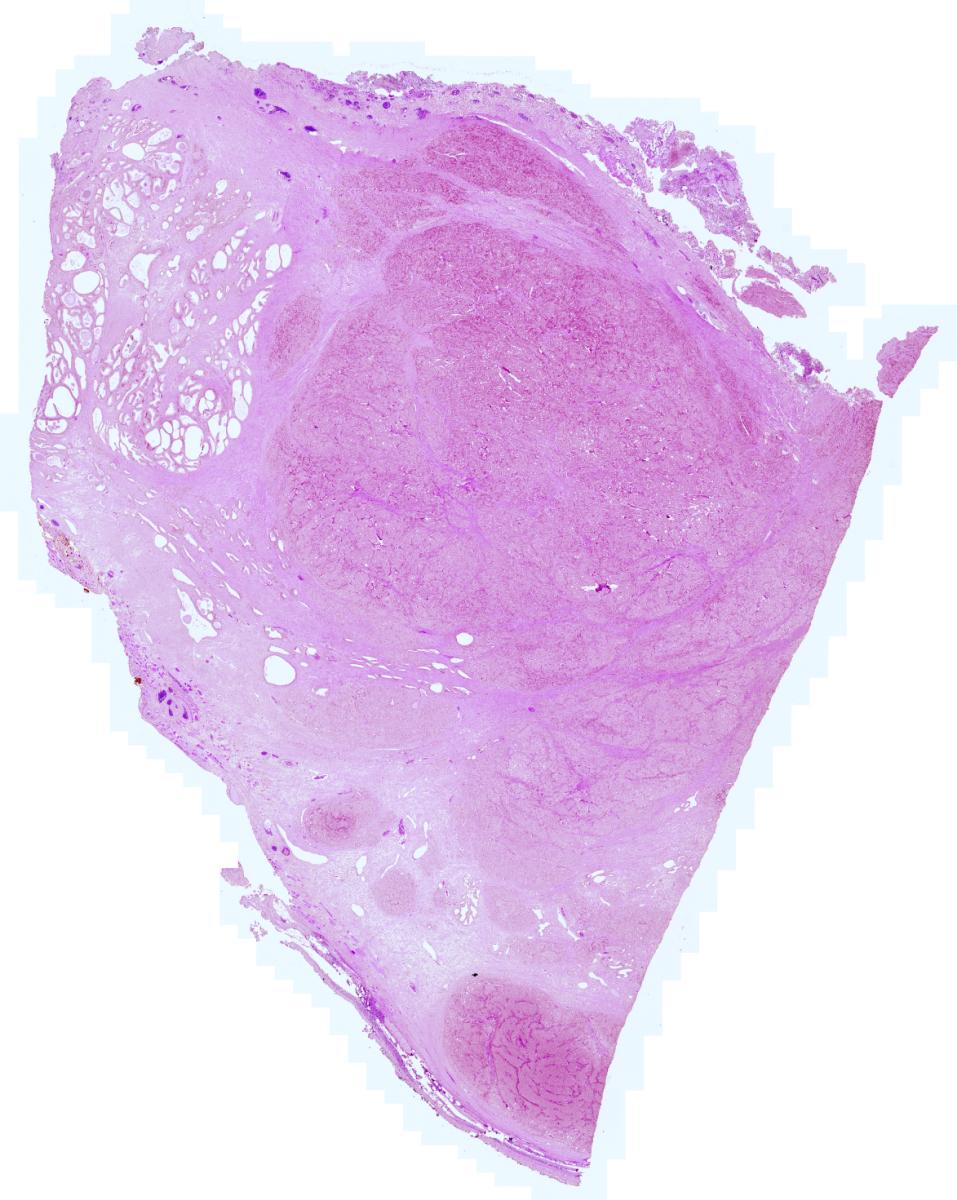}
  \includegraphics[width=0.49\textwidth,scale=0.8]{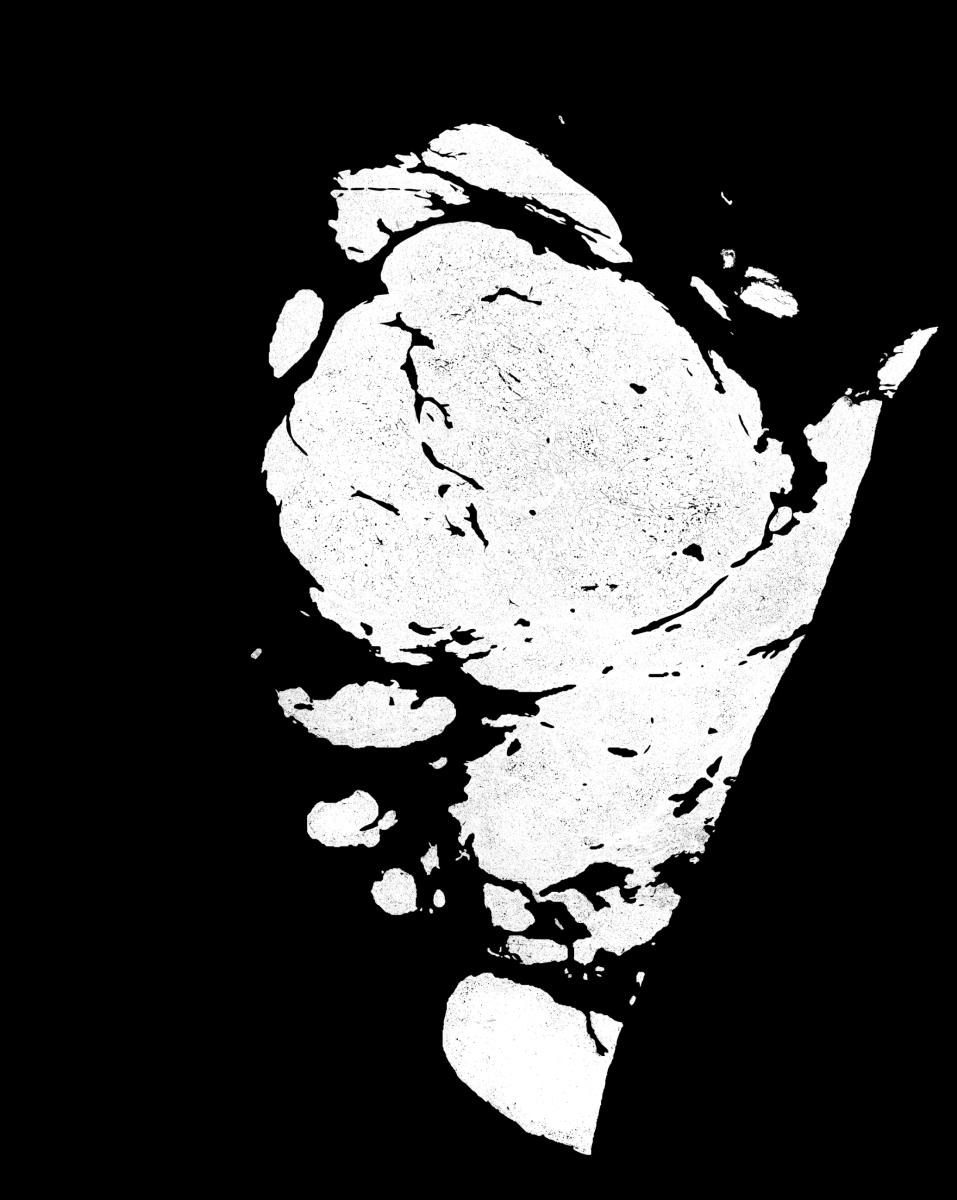}
  \caption{An example of immunofluorescence channels: red - AMACR, green - epithelial channel, blue - basal cells (top left), a WOB heatmap (orange) generated from the channels overlaid on the immunofluorescene channels (top right), a corresponding H\&E staining (bottom left), and a final binary WOB mask (bottom right) for a prostatectomy.}
  \label{example_ma_process}
\end{figure}

\begin{figure}
  \centering
  \includegraphics[width=0.4\textwidth,scale=0.8]{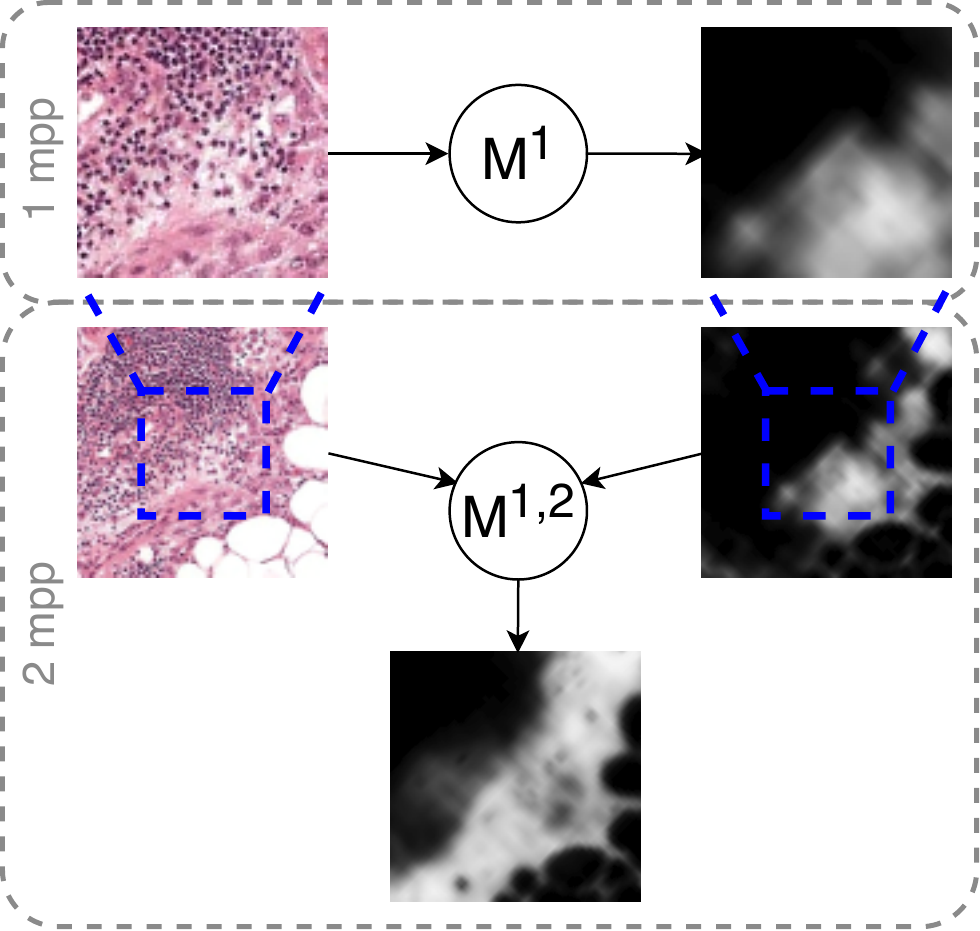}
  \caption{An illustration of our compound model. A first \textbf{unet} $M^1$ is trained on 1 mpp. A second \textbf{unet} $M^{1,2}$ is trained on 2 mpp by taking the output of $M^1$ as an extra channel.}
  \label{fig:compound_model}
\end{figure}

\begin{figure}
  \centering
  \includegraphics[width=1\textwidth,scale=1.0]{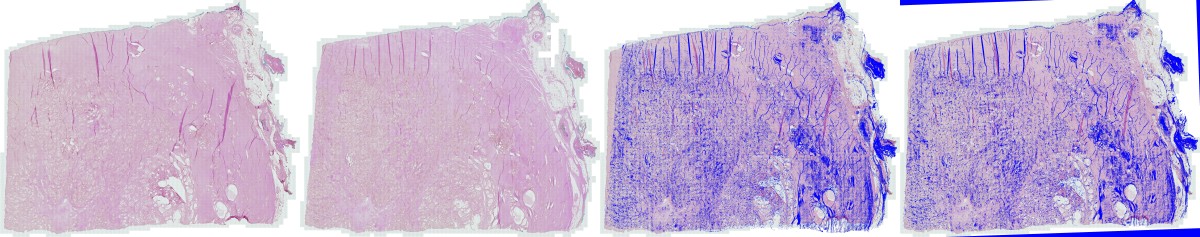}
  \caption{Examples of registration for original and consecutive prostatectomies. The first image is the original image and the second image is the consecutive slice. The third and the fourth images are the original image overlayed with an outline of the consecutive before and after registration.}
  \label{fig:registrations_examples}
\end{figure}

\begin{figure}
  \centering
  \includegraphics[width=1\textwidth,scale=0.8]{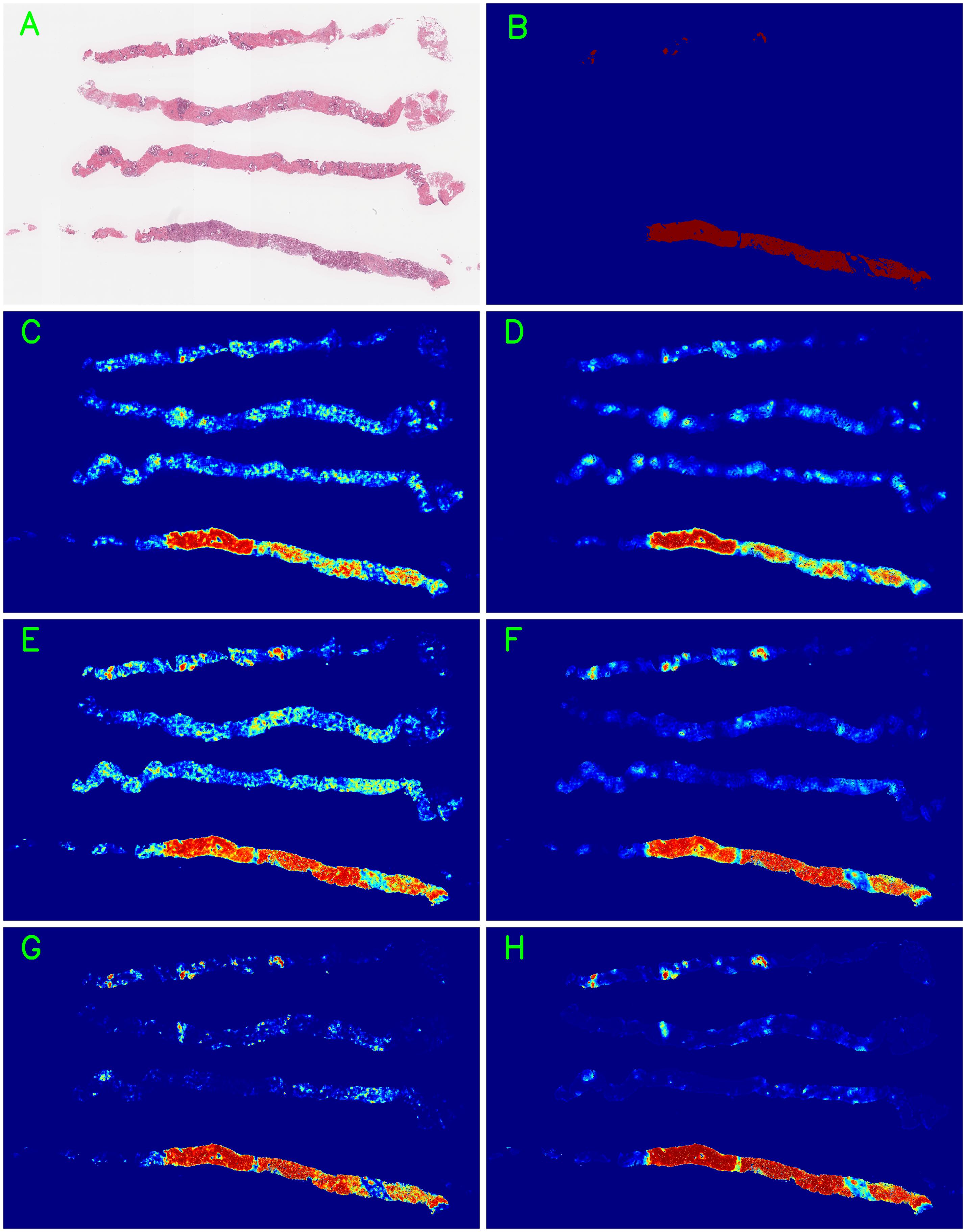}
  \caption{An example of predictions and the ground truth for a biopsy. (A) is an H\&E staining, (B) is an annotation by a pathologist, then there are predictions by (C) $M^{1}_{pr}$, (D) $M^{1,2}_{pr}$, (E) $M^1_{bi}$, (F) $M^{1,2}_{bi}$, (G)  $M^1_{pr,bi}$, and (H) $M^{1,2}_{pr,bi}$.}
  \label{fig:example_biopsy_prediction}
\end{figure}

\begin{figure}
  \centering
  \includegraphics[width=1\textwidth,scale=0.8]{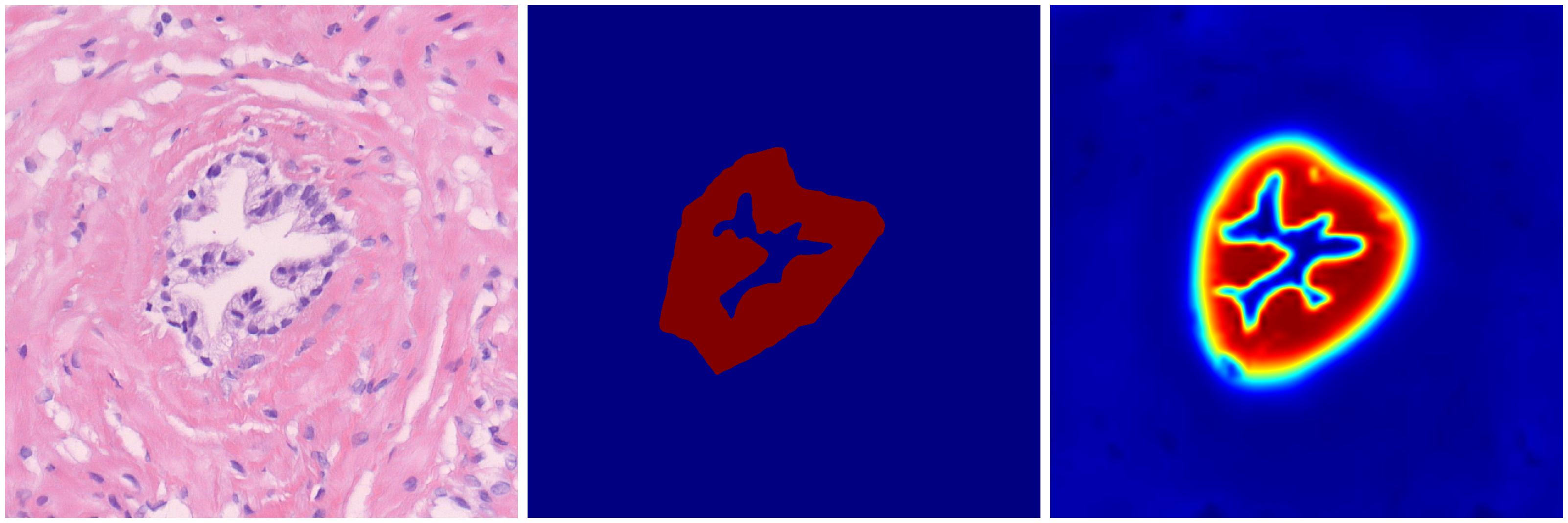}
  \includegraphics[width=1\textwidth,scale=0.8]{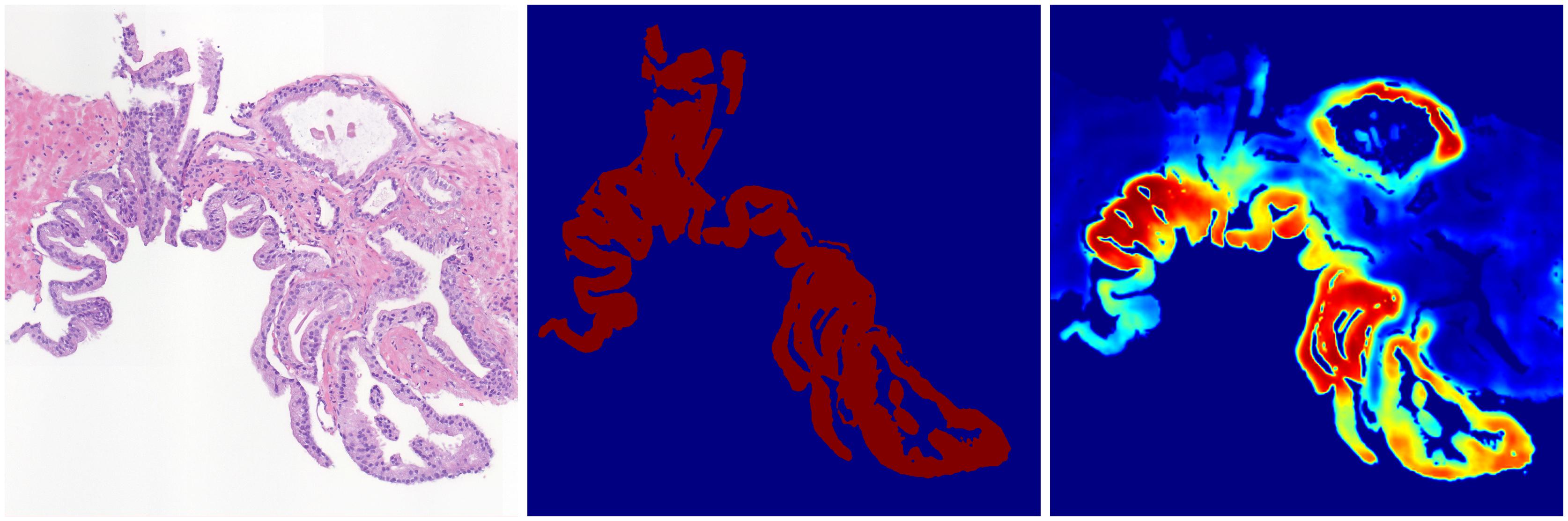}
  \includegraphics[width=1\textwidth,scale=0.8]{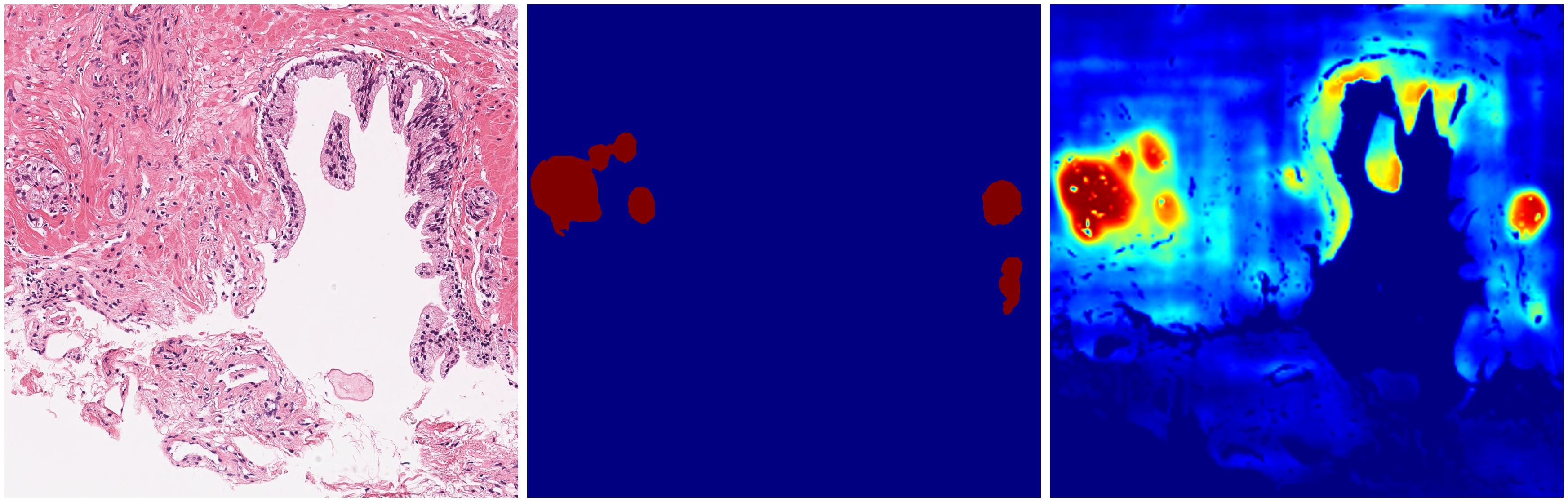}
  \caption{Examples of missed by three pathologists WOB areas from three biopsies (arranged row-wise). In each row, there is an H\&E area first, then a corresponding WOB annotation, and finally a prediction by $M^{1,2}_{pr,bi}$. For the first biopsy, two out of three pathologists totally missed all WOB areas, and one pathologist found WOB areas with Sensitivity of 0.75 and Specificity of 1.00. For the second and the third biopsies, all the three pathologists missed all WOB areas in the biopsies. On contrary, our models detected and accurately outlined most of them.}
  \label{fig:missed_wobs}
\end{figure}

\begin{table}
\centering
\caption{Information on datasets used for training and testing.}
\begin{tabular}{|c|c|c|c|c|c|c|c|c|c|c|c|c|c|c|c|c|c|c|}
\hline \multirow{3}{*}{\#} & \multirow{3}{*}{Type} & \multirow{3}{*}{Scanner} & \multicolumn{7}{c|}{Train} & \multicolumn{7}{c|}{Test} \\
\cline{4-17}
           &      &         &   \multirow{2}{*}{$N_{im}$} & \multicolumn{6}{c|}{Grade Group**}  & \multirow{2}{*}{$N_{im}$} & \multicolumn{6}{c|}{Grade Group**}  \\ 
           \cline{5-10} \cline{12-17}
           &      &         &     & 5 & 4 & 3 & 2 & 1 & 0 & & 5 & 4 & 3 & 2 & 1 & 0 \\ 
\hline
	   \#1 &  prostatect. & A & 40  & 8  &  1 & 12 & 29 & - & - & 10 & 2 & - & 1 & 7 & - & - \\ 
	   \#2 &  prostatect. & A* & 43  & 10 &  1 & 12 & 33 & - & - & 13 & 4 & - & 1 & 8 & - & - \\ 
	   \#3 &  prostatect. & B & 40  & 8  &  1 & 12 & 29 & - & - & 10 & 2 & - & 1 & 7 & - & - \\ 
	   \#4 &  prostatect. & B* & 41  & 11 &  1 & 11 & 32 & - & - & 14 & 4 & - & 1 & 9 & - & - \\  
	   \#5 &  prostatect. & C & 35  & 9  &  1 & 10 & 24 & - & - & 9 & 2 & - & 1 & 6 & - & - \\  
	   \#6 &  prostatect. & C* & 30  & 7  &  1 &  9 & 23 & - & - & 10 & 2 & - & 1 & 7 & - & - \\ 
	   \hline
	       &  Total         &           & 229 & 53 & 6 & 66 & 170 & - & - & 66 & 16 & - & 6 & 44 & - & - \\ 
\hline
       \#7 &  biopsies & D & 21 & \multicolumn{5}{c|}{20} & 1 & 7 & \multicolumn{5}{c|}{7} & -\\  
	   \#8 &  biopsies & E & 97 & \multicolumn{5}{c|}{81} & 16 & 56 & \multicolumn{5}{c|}{38} & 18\\ 
	   \hline
	       &  Total      &           & 118 & \multicolumn{5}{c|}{101} & 17 & 63 & \multicolumn{5}{c|}{45} & 18\\
\hline
\end{tabular}
\label{tab:data_summary}
Prostatectomies were scanned with three different scanners A, B, C and biopsies with D, E.\\
*consecutive slices of tissues; **Gleason Group 0 is for benign samples.
\end{table}

\end{document}